\documentclass[10pt,twocolumn,letterpaper]{article}
\usepackage{wacv}
\usepackage{times}
\usepackage{epsfig}
\usepackage{graphicx}
\usepackage{amsmath}
\usepackage{amssymb}
\usepackage{amsmath,amssymb,amsfonts}
\usepackage{algorithmic}
\usepackage{graphicx}
\usepackage{textcomp}
\usepackage{xcolor}
\def\BibTeX{{\rm B\kern-.05em{\sc i\kern-.025em b}\kern-.08em
    T\kern-.1667em\lower.7ex\hbox{E}\kern-.125emX}}
\wacvfinalcopy 

\def\wacvPaperID{****} 

\ifwacvfinal\fi
\begin{document}

\title{The Sloop System for Individual Animal Identification with Deep Learning\thanks{This work is funded in part by ONR grant N00014-19-1-2273, VTSIX LTD., and the MIT Lincoln Laboratory.}}
\author{Kshitij Bakliwal and Sai Ravela\\
Earth Signals and Systems Group\\
 Massachusetts Institute of Technology\\
77 Massachusetts Avenue, Cambridge MA 02478 USA \\
ravela@mit.edu, http://essg.mit.edu}

\maketitle
\begin{abstract}
The Sloop system indexes and retrieves photographs of non-stationary animal population distributions. To do this, it adaptively represents and matches generic visual feature representations using sparse relevance feedback from experts and crowds. Here, we describe the Sloop system and its application, then compare its approaches to standard deep learning formulations. Results suggest that some of the key ideas that enable Sloop's performance may also benefit deep learning approaches to individual identification. 
\end{abstract}
\thispagestyle{empty}
\section{Introduction}
Effective conservation strategies for rare and endangered species demand unbiased and precise individual life histories~\cite{gambleravela08}. Capture-Mark-Recapture (CMR) studies using photographs enables individual animal tracking. To overcome difficulties with manual photo searches, typically employing {\it ad hoc} markings, computer-based pattern recognition approaches~\cite{kelly01,hiby01,mizroch90,araabi00,arzou05,ravela2004recognizing,gambleravela08,yang09,ravela13} have developed.

  Unfortunately, fully automated recognition machines that would excel on any species simply do not exist. Populations are often non-stationary (e.g. consider insects~\cite{Vyas-Patel034819}), and the demand for high-recall recognition~\cite{gambleravela08} is often significant. Much pre-processing may be necessary to deal with myriad conditions in the wild. These issues make both ``fully automated" approaches and a sole  ``re-identification" focus limiting for numerous conservation questions pertaining to small and perhaps relatively abundant species. 

Taking a retrieval approach, practical solutions today benefit from some degree of human involvement in the identification process. Early approaches demanded too much~\cite{kelly01,hiby01}, but the substantial progress now reliably enables high recall in large image databases. 

Two key ideas have made this possible. The first is to use relevance feedback, accelerated using crowd-sourcing\footnote{We make no distinction here between citizen scientists and crowds.}, to  adapt representations and matchings of image features incrementally and online. We've shown that initially-weak recall quickly improves to very high recall levels~\cite{sloop}. Over time, field experience has shown that self-selection mechanisms effectively leverage crowds, and the amount of human input falls dramatically over feedback iterations~\cite{sloop} — similar ideas applied to pre-processing lead to increased levels of automation and reduced human effort. The second idea is to leverage an ensemble of recognizers, both bagged and boosted, the latter as a cascade of selective methods following invariant ones. 

Sloop is the first system to incorporate the aforementioned ideas for individual animal identification, using which it has consistently shown gains in recall in numerous applications starting from a bag of images with {\bf no} known identities. 

The emergence of learning approaches, particularly end-to-end deep learning, has the potential to advance this paradigm. Could end-to-end learning replace the ``adapt generic features using relevance feedback" paradigm? The comprehensive nature of deep learning and its apparent widespread success is attractive, and the question is essential, to be sure, even when we leave out the detection part of the identification problem as we do in this paper. 

The result of investigating this question, as shown in this paper, suggests that standard deep learning does not yield the desired performance. ''Pre-conditioning" with extracted features appears important to improve neural performance. Further, ideas central to Sloop's success remain, we posit,  necessary for a deep learning approach to succeed.

 The remainder of this paper is organized as follows. In Section~\ref{sec:rw}, precursors and development of Sloop are discussed. In Section~\ref{sec:arch}, the Sloop system architecture, workflow, and methods are discussed. Applications to various species are summarized in Section~\ref{sec:app}. In Section~\ref{sec:learn}, comparisons with deep learning approaches are made on species previously reported. The paper concludes  in Section~\ref{sec:concl}.

\section{Developments Leading To Sloop}
\label{sec:rw}
In late 1996,  interest developed in identifying  ``similar" animals using multiscale differential features~\cite{ravela02}. In 2003, the approach succeeded for recognizing individual salamanders~\cite{ravela2004recognizing};  a clear departure from popular 3D deformable models and ad-hoc methods. By 2007,  local feature methods delivered the first significant biological analyses~\cite{gambleravela08}. The Sloop system was born~\cite{ravela08a,yang09} endowed with local features, and diffeomorphic and scale-cascaded alignment as its base methods. Relevance feedback was then introduced~\cite{ravela02,ravela08a}, as were example-based specularity removal~\cite{runge09} and segmentation methods.  Following a redesign of the Sloop system, by 2012, we incorporated crowdsourced relevance feedback~\cite{ravela13,finn14}, and ingested randomized representations, hybrid contexts into the available routines for creating workflows. Embedded systems versions of Sloop emerged (cellphones, autonomous systems) along with operational use~\cite{sloop} that demonstrated large-scale successes~\cite{otagodailytimes12}. Sloop then became a distributed retrieval system for individual animal identification that reconstructs capture or encounter histories with exceedingly high recall (essential for many conservation questions).

 Incremental online learning in the form of relevance feedback, a key Sloop feature, enables indexing completely unlabeled images, and dealing with non-stationary distributions. Unlike prior approaches~\cite{kuhl2013animal,holmberg2009estimating,town2013manta}, Sloop leverages crowdsourcing for relevance feedback at scale, and with automatic quality control.  Sampling/verification burdens have eased quickly in every application thus far, dramatically improving system recall.  Our experiments suggest that the cumulative advantages of these steps, for example, indexing a $10{,}000$-size collection can, in principle, be accelerated and completed, from scratch, in a few person-days.

 Within the realm of Animal Biometrics, Sloop appears to be the first operational system showing algorithms can help reduce human effort while human feedback can improve system performance. Together, they can produce extensible, scalable, and effective large-scale deployments. The result of deploying Sloop on multiple species has been highly encouraging. It may serve as a useful model to integrate biologists, computational vision researchers, and citizen scientists in a unified framework.

\section{Sloop Architecture}
\label{sec:arch}
\begin{figure}
    \centering
    \includegraphics[width=3.0truein]{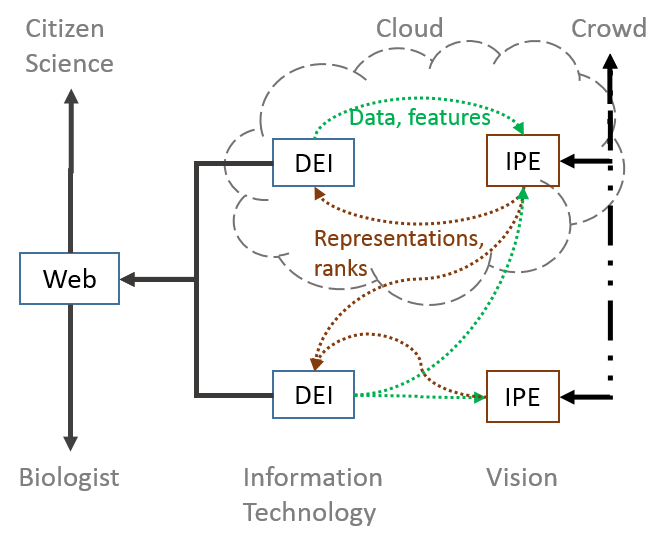}
    \caption{The Sloop system consists of a Data Exchange and Interaction Server (DEI) and an Image Processing Engine (IPE). The user owns the DEI, which manages the image databases and user interactions. The IPE consists of tools and mechanisms to execute workflows, and they interact with DEIs to accomplish identification and indexing tasks. The computation may invoke crowds and now typically executes on AWS. See text.}
    \label{fig:slooparch}
\end{figure}
The Sloop system employs a twin server architecture called the Image Processing Engine (IPE) and Data Exchange and Interaction server (DEI), respectively. This design eases systems integration, quickly absorbing code often provided in Matlab or Python, with suitable wrappers and tests. It also gives the user (biologist) primary control over their data, representations and inferences. Sloop's data access model demands negotiating permissions with the user. As a consequence, the  IPE server requests to provide services.

The DEI manages the databases, including necessary features, representations, scores, and identities. The DEI registers with a name service that the IPE system provides. After that, any IPE begins its work by looking for a DEI, authenticating itself, checking the DEI tables for work to be performed (such as pre-processing, matching, and other steps) and advancing the state of an image in stages from "raw" to eventually "indexed." State advances may require user inputs, which are asynchronously gathered (and in parallel) during the indexing process.  Within a DEI, Sloop provides a full ACID-compliant database with a lightweight Node.js server and a RESTful API over a Postgres backend. The system allows the creation of dynamic indexing workflows, rendered with 3D WebGL, and Angular.js using Bootstrap CSS, and executes on mobile and desktop browsers. 

  Several methods are contained within the IPE (also see~\cite{sloop} for a  description). Pre-processing methods selecting fiducial points, lines, curves, and labeling are available. Mean-shift, graph-cut and CNN-based segmentation, illumination correction techniques~\cite{runge09}, and spline-based  rectification~\cite{ravela2004recognizing} are included.  Multiscale differential feature patches~\cite{gambleravela08}, histograms~\cite{ravela2004recognizing}, all families of local features including affine invariant features~\cite{ravela04b},  hybrid shape contexts, kernel and auto-encoder approaches for dimensionality reduction are included.  Multi-method aggregation, randomized representations, iterated correspondence and RANSAC alignment, and scale-cascaded alignment are also available. 

IPEs and DEIs can co-exist in a virtual machine, as cloud instances, or they can be separate entities. Thus, systems from desktop (SloopLite), mobile, and embedded application (SloopFlyer) to virtual SloopMachines,  and SloopNetworks for enterprise-scale application are available. 

Sloop's  methods are compiled into species-specific workflows, which are finite state machines that begin with image upload and end with the image acquiring an identity. Workflows establish configuration (e.g. views of an animal),  the sequence of pre-processing steps needed, choice of matcher, including fielded or unstructured searchable metadata, and the places where human input is required (see~\cite{sloop}).

In typical practice, the user consults with a (vision) expert on the Sloop team (in contrast to "download and run" approaches) to prototype a workflow on a few images. After that, the system is run in a ``sandbox'' mode for  indexing an initial collection (often unlabeled), and may require human verification. After initialization, the user may use our IPEs during operation, or they may share their own IPE installation with the Sloop community.  Once the initial collection is indexed, assimilating new images involves comparing them with existing cohorts or singletons to maintain updated capture histories. In practice,  a few top matches are verified (typically $1\%$ of the indexing pool) through crowdsourcing with automatic crowd-selection procedures, see~\cite{sloop}.
\section{Application}
\label{sec:app}
The Sloop system application is reaching many species. In the list that follows, the initial species and their AUC performance without and with relevance feedback is presented (in parentheses): jewelled Gecko ({\it Naultinus gemmeus}) ($90\%$, $99\%$), grand and Otago skink ({\it Oligosoma grande} ($96\%$, $99\%$) (and {\it Oligosoma otagense}) ($91\%$, $99\%$) whale shark ({\it Rhincodon typus}) ($94\%$, $97\%$), Fowler's toad ({\it Anaxyrus fowleri}) ($91\%$, $98\%$) tiger salamander ({\it Ambystoma tigrinum}) ($90\%$, $98\%$)and marbled salamander ({\it Ambystoma opacum})($93\%$, $98\%$). 
\begin{figure}
    \centering
    \includegraphics[width=3.1truein]{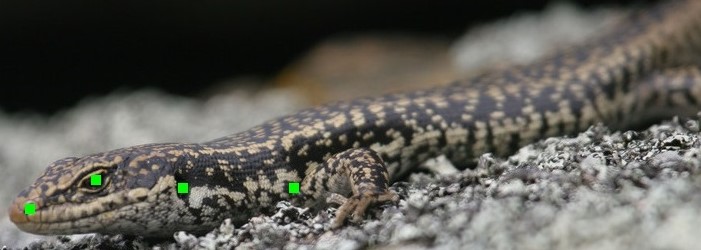}
    \caption{A skink image with body features where patches are extracted.}
    \label{fig:skink}
\end{figure}

In most of these cases, a $97\%$  or above was considered an acceptable standard. Each application used its workflow, which may consist of variations in pre-processing as well as matching methods. In each case, the collection of images was indexed from the first bag of unidentified photographs, and typically two relevance feedback iterations were used. The resulting confidence during field use has been very high~\cite{sloop}. 

\section{Learning Application}
\label{sec:learn}
\begin{figure}
    \centering    
    \includegraphics[width=3.25truein]{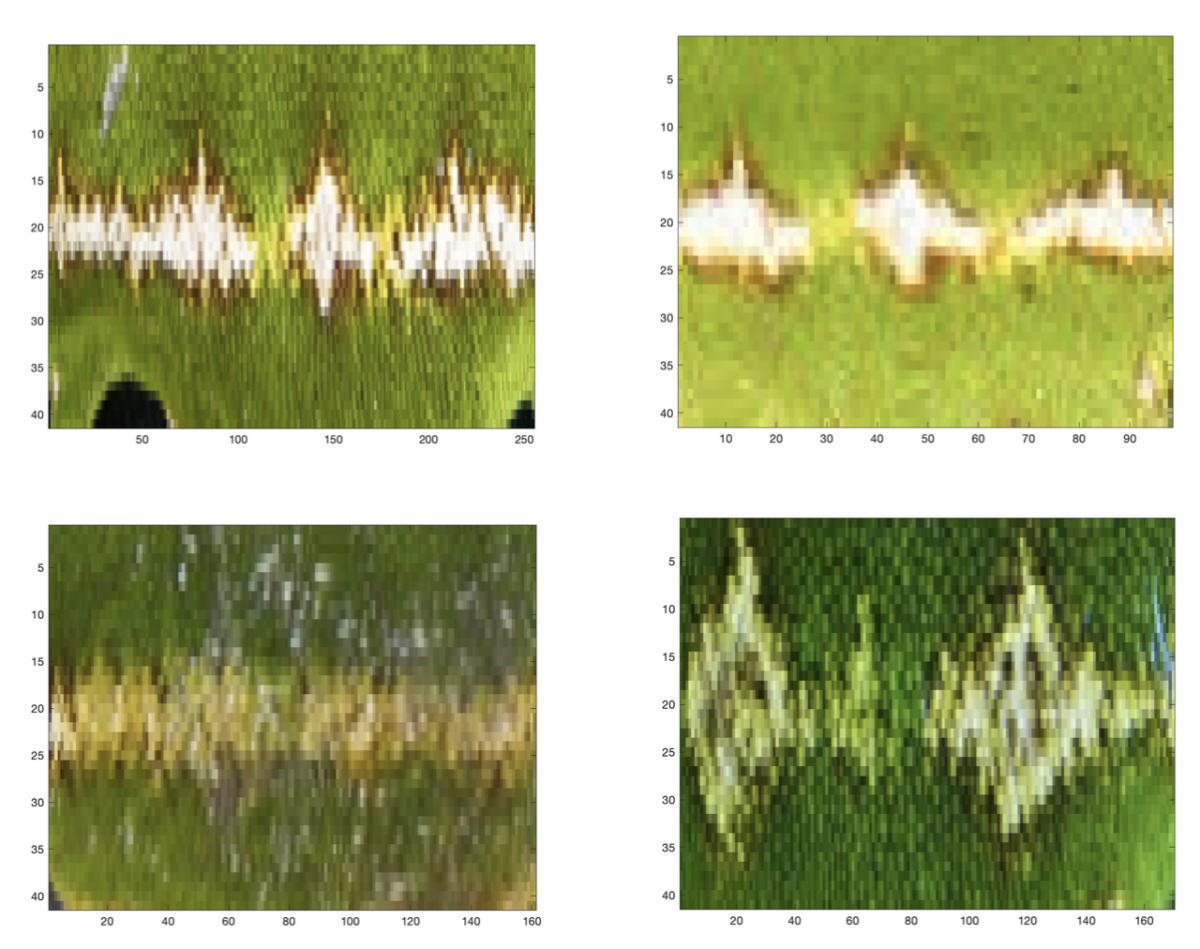}
    \caption{Gecko patch photos post pre-processing. The top two are the same individual, while the bottom two are different. }
    \label{fig:gecko}
\end{figure}

 To advance Sloop, we present two applications of a deep learning approach, one applied to both species of skinks (see Figure~\ref{fig:skink}) and the other applied to the jeweled Gecko (see Figure~\ref{fig:gecko}. Duyck et al.~\cite{sloop} present the prior results and details of the data pre-processing steps). 

In the first set of experiments with the Skink dataset, we employ a Siamese twin model~\cite{siamese} using the pre-trained Alexnet~\cite{alexnet} as the base network, then stripping off the last layer to obtain features. 
A Support Vector Machine with a radial basis function kernel classifies the feature vector difference. The results in a $2/3, 1/3$ training-test split showed very high precision $99\%$, but the recall was quite low $20\%$.

We conducted a second experiment for end-to-end learning with the Gecko dataset, see Figure~\ref{fig:gecko}. Here, we apply a Siamese-twin (similarity-based) model with a triplet loss function for training~\cite{triplet}. 
We use Transfer Learning for our neural network, stripping off the last 3 layers of the base AlexNet architecture. Euclidean distance is the metric of choice. The results from this model (in terms of AUC) became worse: $56\%$

Several standard concerns emerge, such as how to train a bag of unlabeled images, need for extensive training data and pre-training, the optimal network choice, among others.  We've only begun to address these issues. As a start, we improve input features using measures based on differences in amplitude and geometry. Comparing patches pairwise, we use the divergence of deformation field from diffeomorphic alignment~\cite{yang09}, which has shown to be useful for image matching, as the first feature. Using normalized brightness errors as the second feature, and using both features as inputs to a simple three-layer CNN akin to the Siamese version, the results immediately improved to $90\%$ AUC. The Skink dataset improvements are similar. The primed shallow CNN provides excellent skill; using pairwise deformation matching alone only yields a $68\%$ accuracy. 
\section{Conclusion}
\label{sec:concl}
Sloop delivers high recall animal biometrics as a  coupled human-machine system using sparse relevance feedback. In practice, Sloop is used for non-stationary populations successfully, often indexed from ``scratch." 

When compared with a standard deep learning framework, we notice that base performance is not substantially improved, and the improvements obtained were primarily due to improved handling of {\em a priori} features; here the deformation and brightness errors. Very shallow CNNs were needed. Relevance feedback may still be essential to lift the performance to acceptable user levels, just as for the classical approach, but it is as yet unclear how to recursively incorporate it for realtime use.
{\small \bibliographystyle{ieee}
\bibliography{biblio}}
\end{document}